\documentclass[final]{l4dc2024}

\title[Adaptive Teaching in Heterogeneous Agents]{Adaptive Teaching in Heterogeneous Agents: Balancing Surprise in Sparse Reward Scenarios}
\usepackage{times}
\usepackage{listings}
\usepackage{float}
\usepackage{caption}
\usepackage{lipsum}  % just for placeholder code
\usepackage{xcolor}
\usepackage{wrapfig}

\usepackage{url}
\usepackage{xcolor}
\definecolor{mycitecolor}{RGB}{71, 191, 38}
\definecolor{mylinkcolor}{RGB}{40, 115, 201}
\usepackage{hyperref}
\hypersetup{
  colorlinks=true,
  citecolor=blue,
  linkcolor=blue,
  urlcolor=mycitecolor,
}

\author{
 \Name{Emma Clark\textsuperscript{*}} \Email{emmac4@illinois.edu}\\
 \addr University of Illinois at Urbana-Champaign, Urbana, IL, USA \\
 \Name{Kanghyun Ryu\textsuperscript{*}} \Email{kanghyun.ryu@berkeley.edu}\\
 \addr University of California Berkeley, Berkeley, CA, USA \\
 \Name{Negar Mehr} \Email{negar@berkeley.edu}\\
 \addr University of California Berkeley, Berkeley, CA, USA \\
 \addr \textsuperscript{*}These authors contributed equally to this work.
}

\makeatletter
\def\blfootnote{\xdef\@thefnmark{}\@footnotetext}
\makeatother

\begin{document}

\maketitle
\vspace{-4mm}

\begin{abstract}%
 Learning from Demonstration (LfD) can be an efficient way to train systems with analogous agents by enabling ``Student'' agents to learn from the demonstrations of the most experienced ``Teacher'' agent, instead of training their policy in parallel. However, when there are discrepancies in agent capabilities, such as divergent actuator power or joint angle constraints, naively replicating demonstrations that are out of bounds for the Student's capability can limit efficient learning. We present a Teacher-Student learning framework specifically tailored to address the challenge of heterogeneity between the Teacher and Student agents. Our framework is based on the concept of ``surprise'', inspired by its application in exploration incentivization in sparse-reward environments. Surprise is repurposed to enable the Teacher to detect and adapt to differences between itself and the Student. By focusing on maximizing its surprise in response to the environment while concurrently minimizing the Student's surprise in response to the demonstrations, the Teacher agent can effectively tailor its demonstrations to the Student's specific capabilities and constraints. We validate our method by demonstrating improvements in the Student's learning in control tasks within sparse-reward environments\blfootnote{The code is available at \href{https://github.com/labicon/Surprise_based_Teaching}{https://github.com/labicon/Surprise\_based\_Teaching}}.
\end{abstract}


\vspace{-3mm}
\section{Introduction}
Learning from Demonstration (LfD) enables an agent to learn new tasks by imitating another agent. Compared to traditional Reinforcement Learning (RL), LfD is particularly beneficial for learning tasks that require numerous interactions. This benefit is further amplified in multi-agent scenarios, such as assembly~\citep{knepper2013ikeabot} or warehouse systems~\citep{dusadeerungsikul2022multi}. In these situations where agents often share common goals, training multiple agents from scratch for the same task is data-inefficient~\citep{da2017simultaneously}. Hence, utilizing a Teacher-Student framework, where one \emph{Teacher} agent explores the environment and instructs others, can be a more effective approach~\citep{ilhan2019teaching}. For example, consider different robot manipulators on an assembly line. As they share many elements, such as task objectives and working environments, having one robot learn the task first and then instruct others can be a more efficient strategy compared to having multiple robots learn in parallel.

However, agents in these scenarios cannot always be assumed to be entirely identical~\citep{moreira2015robust}. Even in systems with analogous agents, small variations in dynamics can occur between agents due to several reasons, such as under-performing components or different mechanical parts~\citep{da2020agents}. When Student agents learn from a Teacher agent's demonstrations, these discrepancies can lead to performance issues if the Student cannot replicate the Teacher's demonstrations~\citep{ravichandar2020recent}. For instance, in the example of manipulators on an assembly line, variations such as maximum joint angles in different robot models may exist. Then, the Teacher agent must provide demonstrations that are achievable for the Student during the teaching, even when those differences are not explicitly stated. These differences can be inferred through the trajectory of each agent. For example, when teaching a robot arm, the Teacher agent can infer the Student robot's capabilities by observing the robot's maneuvers and accordingly offer demonstrations that align with the Student's movements.

In this work, we address this challenge by presenting a Teacher-Student learning framework that adapts the Teacher's demonstration trajectories to the heterogeneity between the Student and Teacher agents. To quantify this heterogeneity, we introduce the concept of \emph{surprise}, which measures the informational differences between agents or environments. Consequently, the surprise should be small for state-action pairs that have been encountered before and large for those that are unfamiliar. While existing work has focused on surprise between the agent and environment~\citep{berseth2019smirl}, we use surprise to measure the differences between the Teacher's and Student's experiences, thereby quantifying their heterogeneity without explicit knowledge of these differences.
%as the KL-divergence between the base transition probability function and the learned transition probability function~\citep{achiam2017surprise}. Consequently, the surprise will be small for state-action pairs that have been encountered before and large for those that are unfamiliar.

Following~\cite{achiam2017surprise}, we define surprise as the KL-divergence between two transition probability functions. First, we maximize the Teacher's surprise with respect to the environment to incentivize exploration, thus aiding the Teacher in learning its task in sparse-reward environments. Second, for the Student to effectively learn from the Teacher, we argue that the Teacher should also be able to reason about the Student's learning capabilities. Specifically, \emph{we propose that the Teacher must consider the surprise as perceived from the perspective of the Student}. To achieve this, the Teacher learns its own policy with the objective of maximizing its own surprise while minimizing that of the Student. When the Teacher's demonstrations contain state-action sequences different from the Student's dynamics or constraints, the Student's surprise will be large. Consider the example of a manipulation task, where the Student agent has a smaller maximum joint angle than the Teacher. The Teacher may demonstrate trajectories that reach its maximum joint angle; however, the Student will never be able to directly match these trajectories due to physical limitations. Since the Student will never experience such a state, it will perceive a large surprise for such demonstrations. Consequently, the Teacher must adjust its policy to accommodate trajectories within the Student's maximum joint angle.

Our contributions can be summarized in three ways. First, we introduce a Teacher-Student framework in which the Teacher learns its own task in a sparse-reward environment while simultaneously teaching a Student with differing dynamics or constraints. Second, we leverage the notion of surprise, defined as the KL-divergence between the base transition probability function and the learned transition probability function, to enable the Teacher to manage the differing dynamics or constraints of Students without explicit knowledge of these factors. Finally, we empirically demonstrate that our Teacher can adapt its demonstrations to align with the Student's capabilities, resulting in the Student achieving higher rewards.

% In summary, our contribution is as follows:
% \begin{itemize}
%     \item We introduce a Teacher-Student framework in which the Teacher learns its own task in a sparse-reward environment while simultaneously teaching a Student with different dynamics or constraints.
%     \item We employ the concept of \emph{surprise}, defined by the KL-divergence between the base transition probability function and the learned transition probability function, to enable the Teacher to handle the different dynamics or constraints of Students without explicit knowledge of these factors.
%     \item We empirically demonstrate that our Teacher can adapt its demonstrations to match the Student's capabilities, resulting in the Student achieving higher rewards.
% \end{itemize}
\vspace{-2mm}
\section{Related Works}

\noindent \textbf{Teaching Algorithms.} The transfer of knowledge or skills from one agent to another, and the successful utilization of the learned policy by the latter, can be highly beneficial. Algorithms within the family of imitation learning~\citep{hussein2017imitation} and behavior cloning~\citep{9117169} enable an agent to learn a policy from the demonstrations of an expert agent or a human. However, these algorithms typically rely on the expert providing an optimal trajectory and are often limited to scenarios where the teaching and learning agents have similar kinematic systems~\citep{ravichandar2020recent}.

On the other hand, teaching mechanisms places greater emphasis on the Teacher agent’s ability to suggest beneficial actions for the Student~\citep{zimmer2014teacher}, rather than just demonstrating optimal trajectory. Importance advising~\citep{torrey2013teaching} enables the Teacher to calculate a metric to offer advice when beneficial and to correct mistakes. This method allows for advice to be exchanged between simultaneously learning agents even without an expert Teacher; however, its application is primarily limited to discrete action spaces~\citep{da2017simultaneously}. 

In real-world settings, teaching strategies need to adapt to challenges such as differences in kinematics and capabilities between agents, as well as restrictions on agent actions. While some adaptive control models~\citep{petersen2000minimax, nishimura2021rat} can handle differences in noise distributions, they fall short when it comes to differences in dynamics parameters or state constraints. \cite{liu2019state} accounts for different dynamics using state-alignment while neglecting demonstration action. However, they neglect the non-feasible problem where the demonstration trajectory have infeasible state for the learning agent. While \cite{cao2021learning} suggested a feasibility metric for an imitation learning agent focusing on informative teaching data that demonstrates feasible trajectories, we concentrate on the Teacher's side to provide beneficial demonstration data to meet the needs of the Student.
%Even robotic systems that appear equivalent can exhibit small differences in capabilities due to technological limitations~\citep{da2020agents}.

% In real-world settings, teaching strategies need to adapt to challenges such as differences in the kinematics and capabilities between agents or restrictions on agent actions. Even robotic systems that appear equivalent can exhibit small differences in capabilities due to technological limitations~\citep{da2020agents}. While previously discussed methods assume homogeneity between the Teacher and Student agents, we present a framework that is explicitly tailored to situations where the Teacher and Student agent are not identical~\citep{zhang2021confidence, cao2021learning}.  In our method, the Teacher can take into account the Student's dynamics and constraints without needing explicit knowledge of these differences, and adapt its demonstrations to the needs of the Student. 

\noindent \textbf{Surprise in Reinforcement Learning.} Surprise is a concept originating from information theory and derived from Shannon entropy~\citep{shannon2001mathematical}. Surprise serves as a valuable tool for stabilization or exploration in RL. Minimizing surprise drives agents toward familiar state-action pairs~\citep{berseth2019smirl}, while maximizing surprise encourages exploration toward unfamiliar state-action pairs~\citep{mazzaglia2022curiosity}. Bayesian surprise~\citep{itti2009bayesian} has been used for exploration incentive, using latent dynamics models to maximize information gain~\citep{mazzaglia2021self, sun2011planning} or using Bayesian neural networks to model dynamics and maximize surprise~\citep{houthooft2016vime}. However, these methods can be computationally intensive, may not generalize well to continuous action spaces, or may struggle to scale in higher dimensional environments. To overcome these challenges, \cite{achiam2017surprise} defined surprise as the KL-divergence between the learned and true transition probability functions, aiming to motivate exploration with a reduced computational burden.

%Surprise is a useful tool for stabilization or exploration in RL. \cite{berseth2019smirl} includes surprise as a cost to incentivize an agent to return to more familiar state-action sequences, which is useful in high entropy, unstable, or dynamic environments. On the opposite side of this spectrum, surprise maximization during training can drive an agent to explore previously unknown states and actions. \cite{itti2009bayesian} proposed Bayesian surprise which captures the distance between prior belief distribution and posterior belief distribution when acquiring new observations. This concept has been used for exploration incentive, using latent dynamics models to maximize information gain between state transitions~\cite{mazzaglia2021self, mazzaglia2022curiosity, sun2011planning} or by modeling dynamics with Bayesian neural networks and maximizing surprise between transitions~\cite{houthooft2016vime}. However, these methods can be computationally expensive, may not generalize to continuous action spaces, or may not scale well to higher dimensional environments. \cite{achiam2017surprise} proposed surprise defined by KL-divergence of a true transition probability function and learned transition probability function to motivate exploration.

We use the notion of surprise defined in~\cite{achiam2017surprise} to enable the Teacher to demonstrate state-action trajectories that are admissible for the Student. Our goal is to design a teaching method that allows the Teacher to provide effective demonstrations for a Student agent, even in the presence of differing state-space constraints and dynamics.
\vspace{-2mm}

\section{Utilizing Surprise to Instruct Heterogeneous Students}

We introduce a Teacher-Student framework where the Teacher and Student do not necessarily share the same dynamics or constraints. Adapting to these differences is crucial for the Student to learn effectively. Both agents are assumed to lack prior knowledge of the environment, thereby learning simultaneously. We consider sparse reward environments for both the Teacher and the Student. As the Teacher must provide expert demonstrations for the Student to follow, it is necessary that the Teacher first explores the environment to learn a policy that addresses their task. Following~\cite{achiam2017surprise}, we define the Teacher's surprise as the KL-divergence between the Teacher’s learned transition probability model and the true transition probability model of the Teacher’s environment. Since an agent updates its learned model based on training experiences, the KL-divergence between the learned model and the true environment will be small in state-action pairs that have been frequently visited. Consequently, augmenting the Teacher’s reward to be a function of surprise can encourage the exploration of unfamiliar state-action pairs.

Additionally, we propose a strategy for the Teacher that considers the surprise perceived by the Student during the learning process. We define the Student's surprise using \emph{the KL-divergence between the transition probability functions learned by the Teacher and the Student}. If the Student cannot replicate the Teacher's transition probability functions due to differences in dynamics or constraints, this will result in a high value of surprise for the Student. Therefore, by penalizing the Teacher’s reward based on the Student’s \emph{perceived} surprise, we enable the Teacher to account for differences in environments and guide the creation of trajectories that satisfy the constraints or dynamics of the Student.

% \begin{figure}[h]
\begin{wrapfigure}{r}{.65\textwidth}
    \centering
    \includegraphics[width = .65\textwidth]{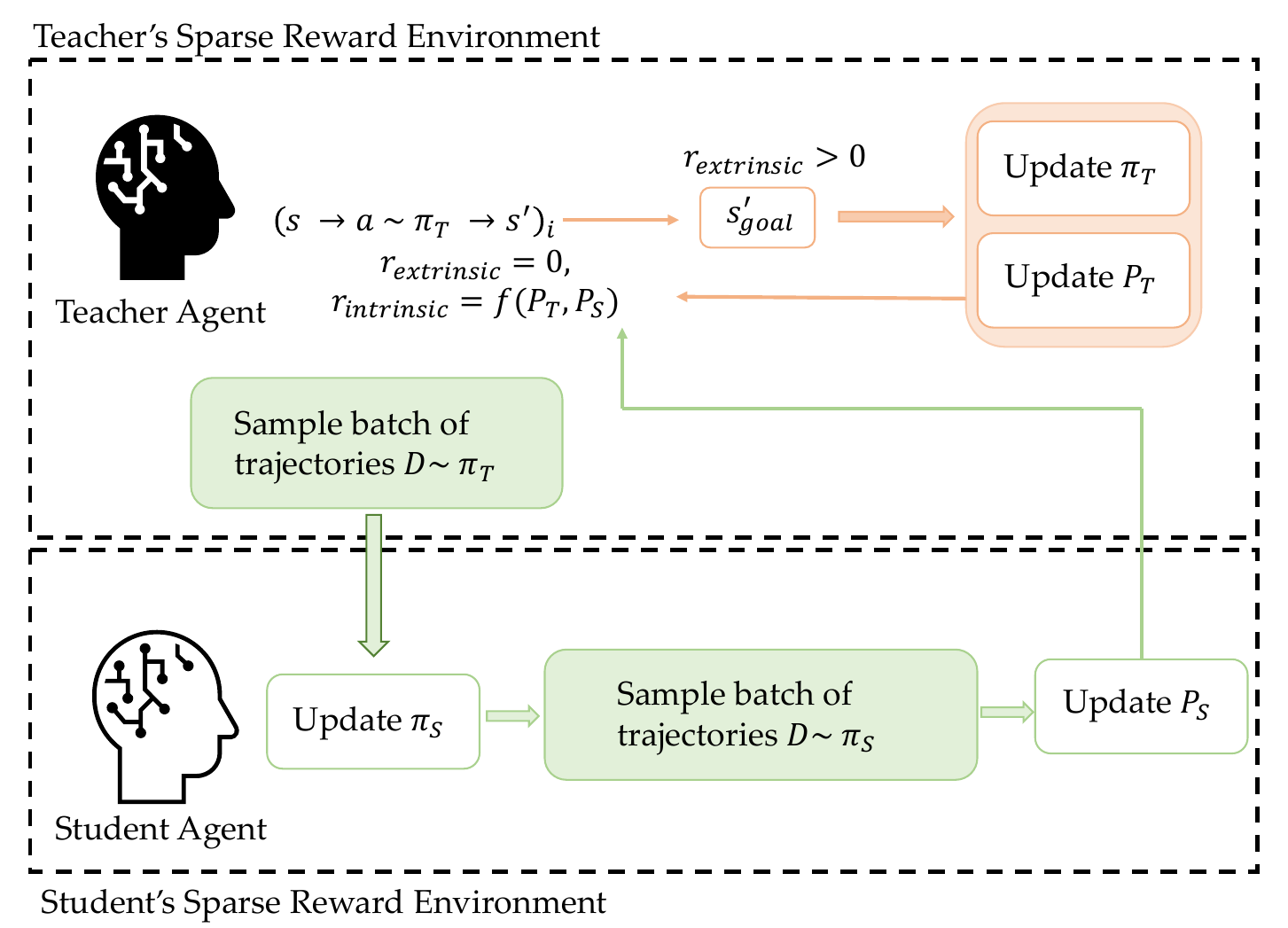}
    \caption{\small Overview of our Teacher-Student framework }%. Two agents acting in their own sparse reward environments, where there is heterogeneity among agents' capabilities or dynamics. The Teacher learns a policy that maximizes its combined extrinsic and intrinsic reward. We shape the intrinsic reward of the Teacher such that it maximizes its surprise while minimizing the Student's surprise; computed using the Teacher and Student's learned probability transition functions. This knowledge allows the Teacher to infer differences between the two environments and adapt its trajectories accordingly.}
    \label{fig:diagram}
    \vspace{-5mm}
\end{wrapfigure}
%\end{figure}

\subsection{Preliminaries}

In our problem, we consider the Teacher and Student as separate agents and denote them by subscript $T$ and $S$, respectively. We model our environments as a Markov Decision Process (MDP) and define this MDP as the tuple $<\mathcal{S}, \mathcal{A}, r_{e}, P, \gamma>$ where the Teacher and Student have their own respective state spaces $\mathcal{S}_T, \mathcal{S}_S$, action spaces $\mathcal{A}_T, \mathcal{A}_S$, and extrinsic reward functions $r_{e_T}:\mathcal{S}_T \times \mathcal{A}_T \rightarrow \mathbb{R}, r_{e_S}:\mathcal{S}_S \times \mathcal{A}_S \rightarrow \mathbb{R}$. Each environment has a true transition probability function $P_T({s}'| s, a)$ and $P_S({s}'| s, a)$, which give the true probability of transition to state $s'$ from the current state $s$ given action $a$. While these transition probability functions depend only on the Teacher's and Student's state-action pairs, respectively, we omit their subscripts for easier notation. The agents have their learned transition probability functions denoted by a subscript $\phi$: $P_{\phi_T}$ for the Teacher and $P_{\phi_S}$ for the Student. We define a stochastic policy for the Teacher $\pi_T(\cdot | s): \mathcal{S}_T \rightarrow \mathcal{A}_T$ and Student $\pi_S(\cdot | s): \mathcal{S}_S \rightarrow \mathcal{A}_S$ as a distribution over possible actions given a state $s$. We assume that the Teacher has full access to the Student's learned transition probability function $P_{\phi_S}$.

\subsection{Surprise for Exploration}\label{sec:exploration_surprise}
 
In a sparse-reward setting, we motivate the Teacher to maximize its own surprise to explore the environment. We assume the Teacher does not have access to $P_T$. Instead, the Teacher utilizes its own learned transition probability function, $P_{\phi_T}$. As the Teacher agent explores the environment, this learned function should converge to the true transition function. Therefore, we can represent the Teacher agent's surprise by the KL-divergence between the learned transition probability function and the true transition probability function.
 \begin{equation*}
     D_{KL}\big(P_T(\cdot|s,a)||P_{\phi_T}(\cdot|s,a)\big) = H\big(P_T(\cdot|s,a),P_{\phi_T}(\cdot|s,a)\big) -H\big(P_T(\cdot|s,a)\big),
 \end{equation*}
where $H\big(P_T(\cdot|s,a)\big)$ is the entropy of $P_T(\cdot|s,a)$ and $H\big(P_T(\cdot|s,a),P_{\phi_T}(\cdot|s,a)\big)$ is the cross entropy of $P_T(\cdot|s,a)$ and $P_{\phi_T}(\cdot|s,a)$.

% , $H\big(P_T(\cdot|s,a)\big)$ is non-finite everywhere. Then, if $H\big(P_T(\cdot|s,a)\big)$ has the same sign everywhere
In deterministic environments with continuous state spaces, $\mathbb{E}_{s,a\sim\pi_{T}} [H\big(P_T(\cdot|s,a)\big)]$ is constant and can be dropped from the optimization problem~\citep{achiam2017surprise}. Therefore, the KL-divergence can be approximated as the cross entropy between the Teacher's learned transition probability function and the environment's true transition probability function. We compute the cross entropy between the two distributions, $P_T(\cdot|s,a)$ and $P_{\phi_T}(\cdot|s,a)$, as
 \begin{align} \label{eqn:teacher_surprise}
    \begin{split}
        H(P_T(\cdot|s,a),P_{\phi_T}(\cdot|s,a)) &=  - \int_{\mathcal{S}} P_T(s' | s,a) \log \big( {P_{\phi_T}(s'|s,a)} \big) ds' \\
        &= \mathbb{E}_{s' \sim P_T(\cdot|s,a)}\big[-\log{P_{\phi_T}(s'|s,a)}\big].
    \end{split}
 \end{align}

\subsection{Computing the Surprise Perceived by the Student} 

In this section, we discuss how the Teacher can tailor trajectories to accommodate a Student with different dynamics or environmental constraints. We define the Student's surprise as a KL-divergence between the Student's learned transition probability function and that of the Teacher. Therefore, the Student surprise for state-action pair $(s,a)$ is
\begin{equation*}
    D_{KL}\big(P_{\phi_T}(\cdot|s,a)||P_{\phi_S}(\cdot|s,a)\big).
\end{equation*}

Because the Teacher has access to both the Student's and Teacher's learned transition probability functions, we can directly calculate the KL-divergence between the two: 
\begin{align} \label{eqn:student_surprise}
\begin{split}
    D_{KL}\big(P_{\phi_T}(\cdot|s,a)||P_{\phi_S}(\cdot|s,a)\big) &= \int_{\mathcal{S}} P_{\phi_T}(s'|s,a) \log \left( \frac{P_{\phi_T}(s'|s,a)}{P_{\phi_S}(s'|s,a)} \right) ds' \\
    &= \mathbb{E}_{s' \sim P_{\phi_T}(\cdot|s,a)}\big[\log{P_{\phi_T}(s'|s,a)} -\log{P_{\phi_S}(s'|s,a)}\big].   
\end{split}
\end{align}

When the Student and Teacher exhibit similar learned transition dynamics for specific state-action pairs, the KL divergence in \eqref{eqn:student_surprise} will be small, resulting in a lower surprise value for the Student. Conversely, if there are significant differences in their learned transition dynamics, stemming from different environment constraints or dynamics, the surprise value perceived by the Student in \eqref{eqn:student_surprise} will be large. In conclusion, a large surprise is likely to occur if the Teacher demonstrates a trajectory that is incompatible with the Student's dynamics or constraints. Therefore, our objective is to minimize this surprise during the Teacher's demonstrations. 

\subsection{Shaping the Teacher's reward for Adaptive Demonstrations}

We define the intrinsic reward $r_{i}$ for the Teacher as a weighted sum of the surprise terms for both the Teacher and the Student. The inclusion of the Teacher’s surprise encourages exploration in a sparse-reward environment, facilitating the exploration of novel state-action pairs. Simultaneously, incorporating the Student’s surprise enables the Teacher to make informed inferences about the Student’s dynamics and constraints. Consequently, this dual consideration allows the Teacher to tailor its demonstrations to align with the Student’s capabilities. Each surprise incentive can be computed using~\eqref{eqn:teacher_surprise} and~\eqref{eqn:student_surprise} as follows
\begin{align} \label{eqn:surprise_bonus}
    \begin{split} 
        &r_{i}(s,a) = \eta_{T}D_{KL}\big(P(\cdot | s,a) || P_{\phi_T}(\cdot | s,a) \big)  - \eta_{S}D_{KL}\big(P_{\phi_T}(\cdot | s,a) || P_{\phi_S}(\cdot | s,a) \big) \\ 
        &\approx \eta_{T}\mathbb{E}_{s' \sim P(\cdot|s,a)}\big[-\log{P_{\phi_T}(s'|s,a)}\big] - \eta_{S}\mathbb{E}_{s' \sim P{\phi_T}(\cdot|s,a)}\big[\log{P_{\phi_T}(s'|s,a)}-\log{P_{\phi_S}(s'|s,a)}\big],
    \end{split}
\end{align}
where $\eta_T$ and $\eta_S$ are the weights of each surprise term. Following~\cite{achiam2017surprise}, $\eta_T$ is given by
\begin{equation}
    \eta_T = \frac{\eta_{0_T}}{\max{\big(1, \frac{1}{|\mathcal{D}|}\sum_{(s,a) \in \mathcal{D}}r_{e_T}(s,a)}\big)},
\end{equation}
where $\eta_{0_T}$ is a predefined constant, $r_{e_T}$ represents the extrinsic return from a trajectory rollout $\mathcal{D}$ of size $|\mathcal{D}|$ following the Teacher's policy. This factor scales the exploration bonus magnitude according to the extrinsic rewards of the environment. The Student's coefficient $\eta_{S}$ is defined analogously, with extrinsic rewards calculated from trajectories sampled from the Student's policy, and the Student's predefined $\eta_{0_S}$ constant.

%For example, if a Teacher agent is receiving large extrinsic reward, the weight on the Teacher's surprise reduces since there is less need for exploration. The Student's coefficient $\eta_{S}$ is defined analogously, with extrinsic rewards calculated from trajectories sampled from the Student's policy, and the Student's predefined $\eta_{0_S}$ constant. In other words, if the Student is receiving a large reward, it reduces the weight on the Student's surprise and focuses on the Teacher's exploration.
The reshaped reward for the Teacher is designed to incentivize exploration of new states while simultaneously disincentivizing visits to states that are excessively unfamiliar to the Student. If the Teacher demonstrates trajectories that either violate the Student's constraints or differ significantly from the Student's dynamics, the Student's surprise will be high. This is due to the Student's learned transition model differing substantially from these trajectories. As a result, in an effort to maximize its reward as outlined in~\eqref{eqn:surprise_bonus}, the Teacher is encouraged to avoid such trajectories and instead demonstrate paths that are aligned with the dynamics and constraints of both the Teacher and the Student.

Finally, the Teacher's objective is to maximize the augmented sum of extrinsic and intrinsic rewards. Therefore, the objective function for the Teacher is as follows
\begin{equation} \label{eqn:max_obj}
    L_T(\pi_T) = \mathbb{E}_{a_t \sim \pi_T(\cdot|s_t)} \big[ \sum_{t=0}^H \gamma^t \big( r_{e_T}(s_t,a_t) + r_{i}(s_t,a_t) \big) \big].
\end{equation}

\subsection{Implementation Details}

We can use any RL algorithm to optimize the Teacher policy that aims to maximize~\eqref{eqn:max_obj}. In our experiments, we implemented Trust Region Policy Optimization (TRPO)~\citep{schulman2015trust} to train the Teacher policy, enabling the examination of both continuous and discrete action spaces. This choice also allows for a direct comparison with~\cite{achiam2017surprise}, which used TRPO as their base optimization method, thus providing a more conclusive analysis of performance differences when incorporating the Student's surprise. For the Student agent, we implement behavioral cloning (BC)~\citep{bain1995framework, ross2010efficient} to learn from the Teacher's demonstration. 

We use probabilistic neural networks~\citep{chua2018deep} to learn the transition probability functions of the Teacher and Student agents. We model the transition probability functions of the Teacher agent as a Gaussian distribution $ P_{\phi_T}(\cdot|s,a) = \mathcal{N}\big(\mu_{\phi_T}(s,a), \Sigma_{\phi_T}(s,a)\big),$ where $\mu_{\phi_T}$ is the learned mean and $\Sigma_{\phi_T}$ is the learned covariance of the distribution. We update the mean and covariance of the distributions by optimizing the negative log-likelihood loss function 
\begin{equation} \label{eqn:loss}
    \text{loss}_{NLL} = \sum_{i = 0}^{N}(\mu_{\phi_T}(s_i,a_i) - s_{i+1}^{T})\Sigma^{-1}_{\phi_T}(s_i,a_i)(\mu_{\phi_T}(s_i,a_i) -s_{i+1})+\log\big({\det \Sigma_{\phi_T}(s_i,a_i)}\big),
\end{equation}
where $N$ is the batch size and $s_{i+1}$ is the sampled true next state of the environment given state and action pairs $(s_i, a_i)$. The same procedures are applied to the Student's learned transition probability function $P_{\phi_S}(\cdot|s,a)$.

At each training epoch, we first update the Teacher's transition probability model with~\eqref{eqn:loss} and the Teacher's policy to maximize~\eqref{eqn:max_obj}. Then, the Teacher gives demonstrations to the Students using the Teacher's policy. We update the Student policy with BC for the Teacher demonstration. Finally, the Student's transition probability model is updated with the trajectory rollout of its policy. 

%\begin{algorithm}
%\Line
%\caption{Teacher-Student Algorithm}\label{alg:cap} \vspace{-2mm}
%\Line \\
%    \While{$i < epochs$}{
%        \For(\tcp*[f]{N = batch size}){$j\gets0,N$}{
%            \For{Teacher}{
%                collect rollouts from $\pi_{T}, \pi_{S}$\; \\
%                update transition probability $P_{\phi_T}(\cdot|s,a)$\; \\
%                optimize reshaped reward with surprisal bonuses %(Equation~\eqref{eqn:surprise_bonus})\; \\
%                update policy $\pi_{T}$ \;
%            }
%            \For{Student}{
%                sample demonstration trajectory from $\pi_{T}$\; \\
%                update policy using behavior cloning $\pi_{S}$\; \\
%                collect rollouts from $\pi_{S}$\; \\
%                update transition probability $P_{\phi_S}(\cdot|s,a)$\;
%            }
%        }
%    }
%\vspace{-2mm}
%\Line
%\end{algorithm}

\section{Experiments}
We implement our algorithm in sparse reward environments introduced by~\cite{houthooft2016vime}: Mountain Car, Cart Pole Swing Up, and sparse Half Cheetah. In Section~\ref{sec:identical_experiments}, we initially test our method with both the Student and Teacher operating in the same environment to ensure that our approach doesn't obstruct the learning capabilities of either agent in homogeneous settings.. Next, in Section~\ref{sec:different_experiments}, we show that our method helps the learning of the Student agent which has a different environment from the Teacher.

\subsection{Identical Teacher-Student Environments} \label{sec:identical_experiments}
\begin{figure}[t]
    \centering
    \subfigure[Mountain Car]{
           \includegraphics[width = 0.315\textwidth]{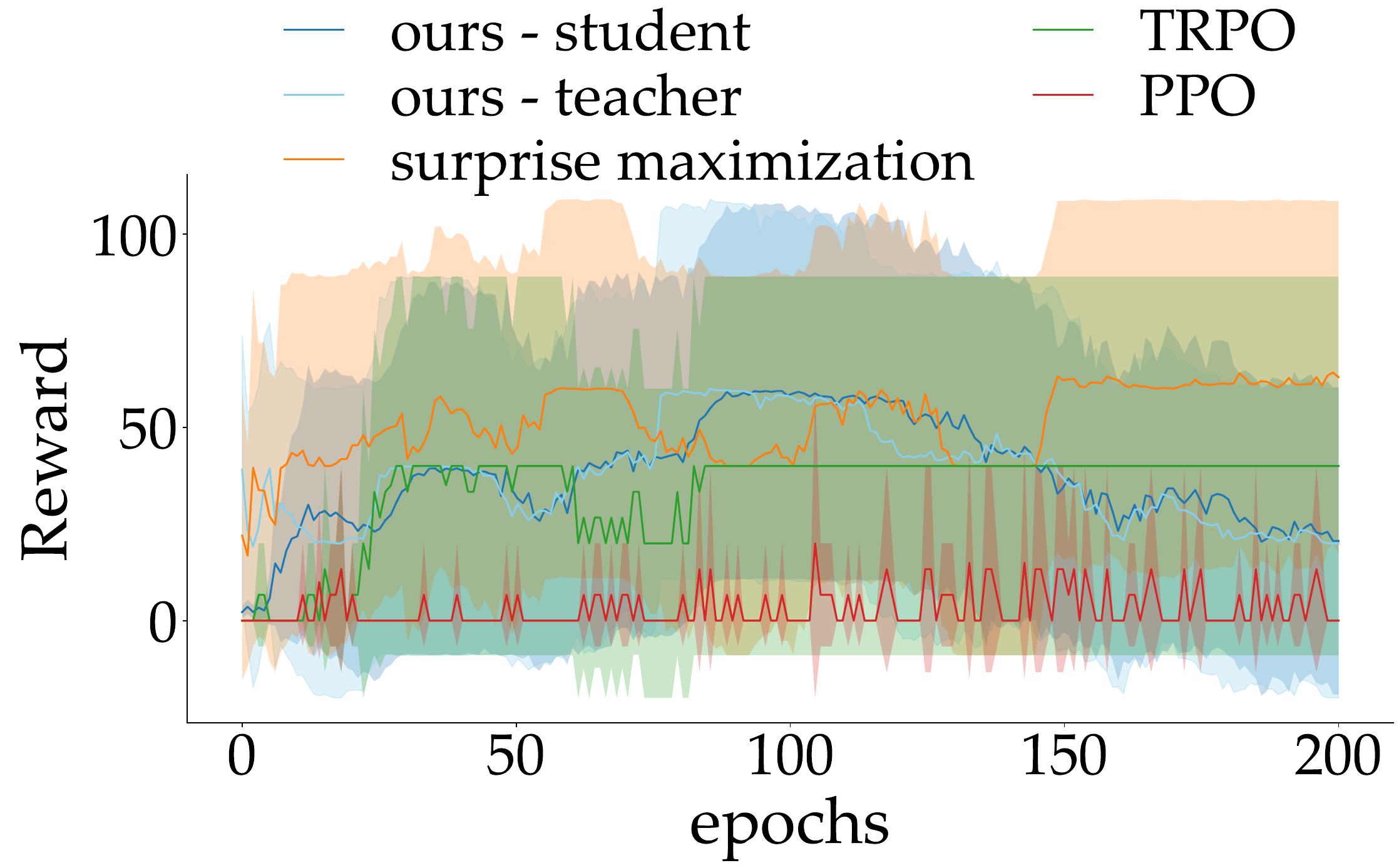}
    }
    \subfigure[Cart Pole Swing Up]{
        \includegraphics[width=0.315\textwidth]{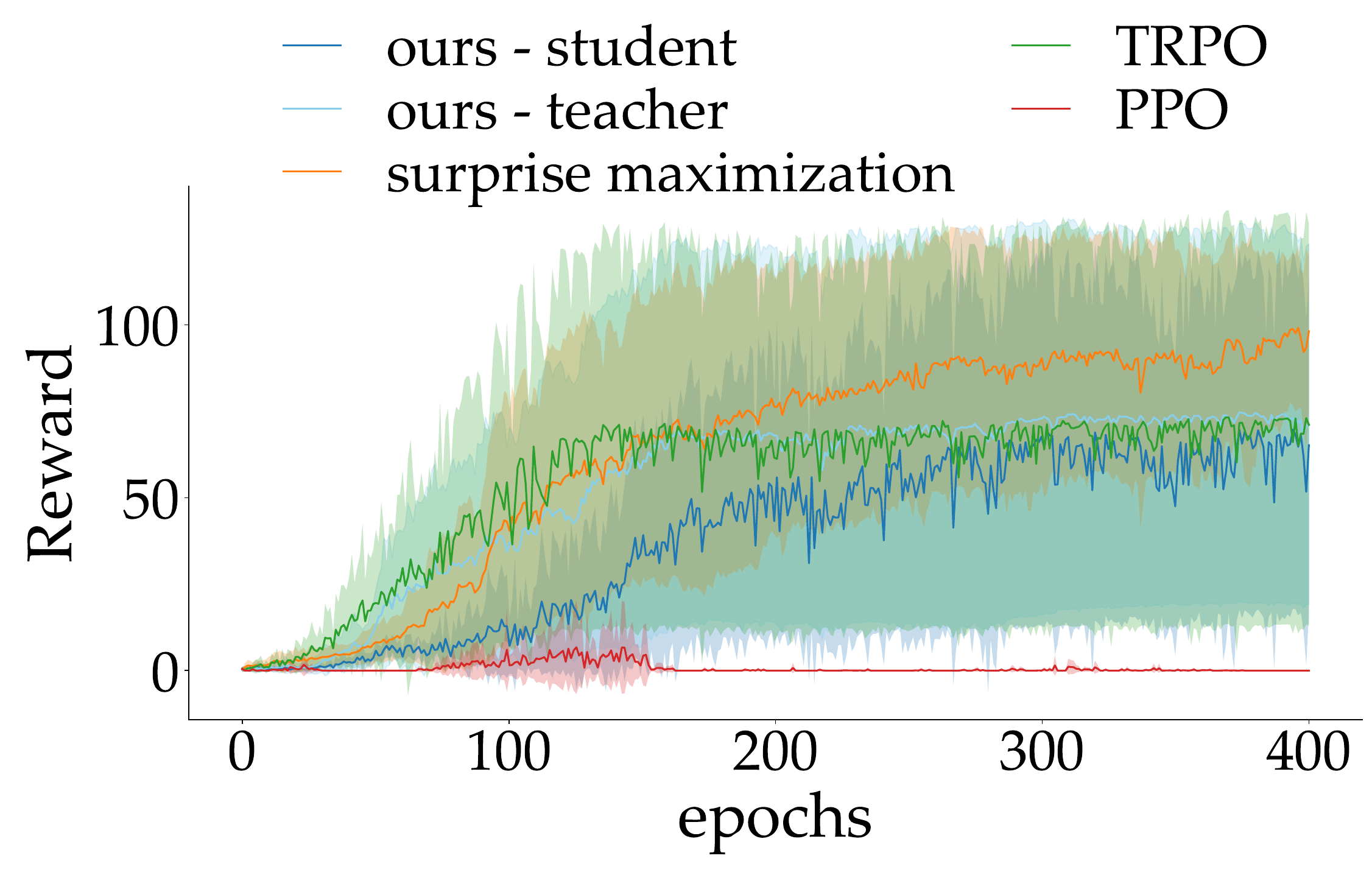}
    }
    \subfigure[Half Cheetah]{
        \includegraphics[width=0.315\textwidth]{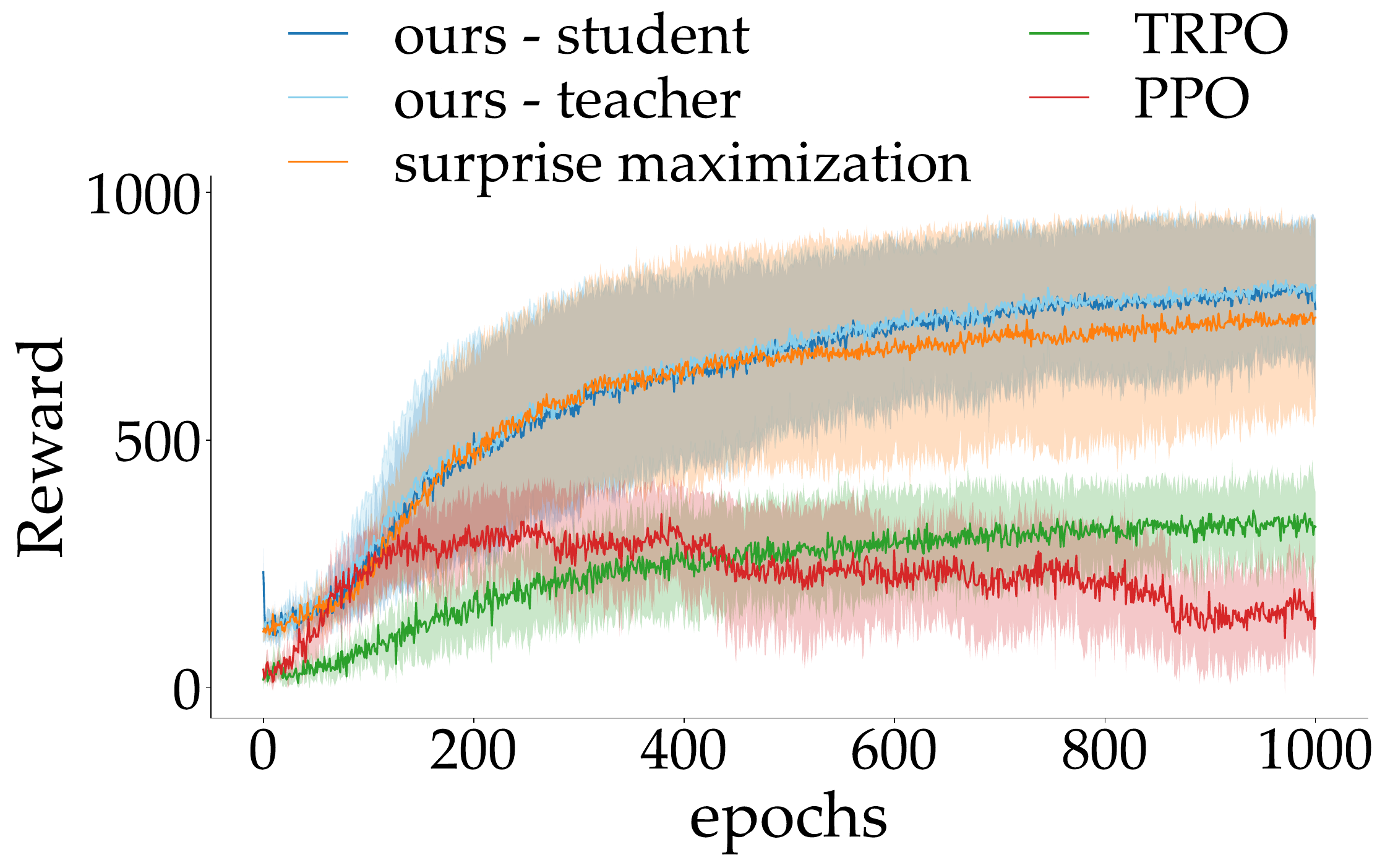}
    }
    \caption{\small
    Mean and standard deviation of reward for the three environments are shown. Results are from 5 random seeds for Mountain Car and Half Cheetah and 8 random seeds for Cart Pole Swing Up. Our algorithm is deployed in a setting where the Teacher and Student are in the same environment. Baselines are trained in a single-agent setting where they are trained without the Teacher. Both Teacher and Student in our Teacher-Student framework can learn successful policy in sparse-reward environments.
    }
    \label{fig:std_env_lc}
    \vspace{-3mm}
\end{figure}
Figure~\ref{fig:std_env_lc} presents the results of our Teacher-Student framework when the Teacher and Student have identical environments. Compared to our Teacher-Student framework, TRPO, PPO~\citep{schulman2017proximal}, and surprise-maximization \citep{achiam2017surprise} algorithms tries to learn a policy without the Teacher. Both Teacher and Student in our method performs at a similar level to that of~\cite{achiam2017surprise}. Therefore, we can conclude that the Student-Teacher framework does not diminish the learning capabilities of the agents. In addition, observing the low returns of TRPO and PPO, we conclude that surprise motivation is necessary for exploration in sparse reward environments.

%Both the Teacher and Student in our method show comparable performance with the baselines in terms of average returns. PPO fails to learn both in the Mountain Car and Cart Pole Swing Up environments. In the Mountain Car environment, TRPO, surprise maximization, and our method have similar learning curves, that show they are learning roughly at the same speed. In the Cart Pole Swing Up environment, our method converges to a similar performance level to TRPO at the end of training, both being outperformed by surprise maximization. In the Half Cheetah environment, PPO and TRPO have similar learning curves, while the surprise maximization method and our method show much better performance than the previous two. Near the end of training, our method starts to diverge from the surprise maximization method, giving a slightly better average return.

%In summary, our method performs at a similar level to~\cite{achiam2017surprise}, when the Teacher and Student have the same environment. In addition, in the case of the Half Cheetah environment, we show slight improvement when considering the average returns. We can see that the Student-Teacher framework does not imperil the learning capabilities of the agents, but even improves them in some cases. 
\subsection{Heterogeneity Between Teacher and Student} \label{sec:different_experiments}
\subsubsection{Experiment Setup}
After confirming that our framework does not impact learning in the homogeneous setting, we test our learning framework in the heterogeneous setting where the Teacher and Student are in environments with different dynamics or constraints where we can examine the ability of the Teacher agent to adjust its trajectories according to the Student's environment. Especially, we focus on settings where the Student's ability is limited compared to the Teacher. This is because even if the Student has superior ability compared to the Teacher, its effectiveness would be limited to the Teacher's policy due to imitating the Teacher's policy.

In the Mountain Car environment, we adjust the car power constant in the dynamics equation. We keep the baseline value of $p = 0.001$ for the Teacher and reduce that of the Student to $p=0.0067$. In this lower-power version, an agent requires a greater force to achieve the desired velocity compared to the higher-power setup. Consequently, the Teacher should demonstrate trajectories utilizing larger force magnitudes than those used in its high-power environment. For the Cart Pole Swing Up environment, we have two experiments with different $x$-position constraints and different pole masses. In the different $x$-position constraints experiment, we have the Teacher's constraint to be $|x_{pos}| \leq 3.6$ while the Student is using $|x_{pos}| \leq 2.4$. For the different pole mass experiments, the Teacher agent has a pole mass of $m=0.1$ while the Student has a heavier pole mass of $m=0.12$. We note that the maximum episode length in the CartPole was increased to $500$ for faster exploration. Finally, in the sparse Half Cheetah environment, the Teacher's head angle is unconstrained while the Student's is constrained to $|\theta_{head}| \leq 1 
 \text{rad}$.

In Figure~\ref{fig:std_env_lc}, we observed that baseline algorithms that lack an exploration motivation failed to learn successful policies in complex sparse-reward environments. Consequently, we used surprise maximization~\citep{achiam2017surprise} as the sole baseline. In this baseline, the Teacher attempts to learn an optimal policy by exploring the environment through surprise-maximization while neglecting the Student's perceived surprise. This comparison highlights the efficacy of incorporating the Student's surprise into the Teacher's reward structure. Meanwhile, our method enables the Teacher to tailor demonstrations more effectively to the Student when their environments are dissimilar. In both scenarios, the Student performs BC on the Teacher's policy. 
\begin{figure}[t]
    \centering
    \subfigure[Mountain Car - Different Power]{
    \begin{minipage}{0.48\textwidth}
        \centering
        \includegraphics[width=0.75\linewidth]{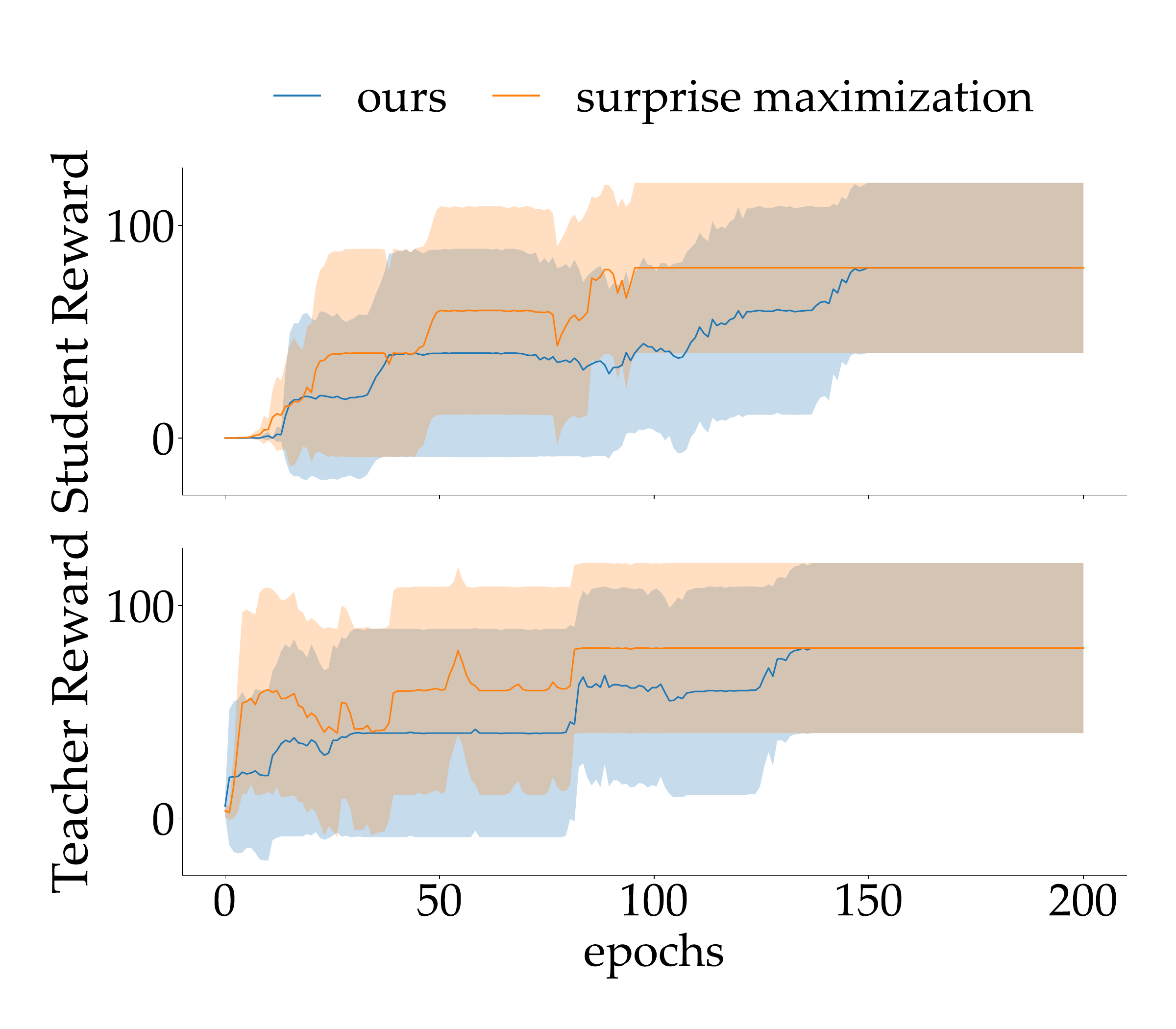}
    \end{minipage}
    }
    %\hspace{1cm}
    \subfigure[Half Cheetah - Constrained Tip Angle]{
    \begin{minipage}{0.48\textwidth}
        \centering
        \includegraphics[width=0.75\linewidth]{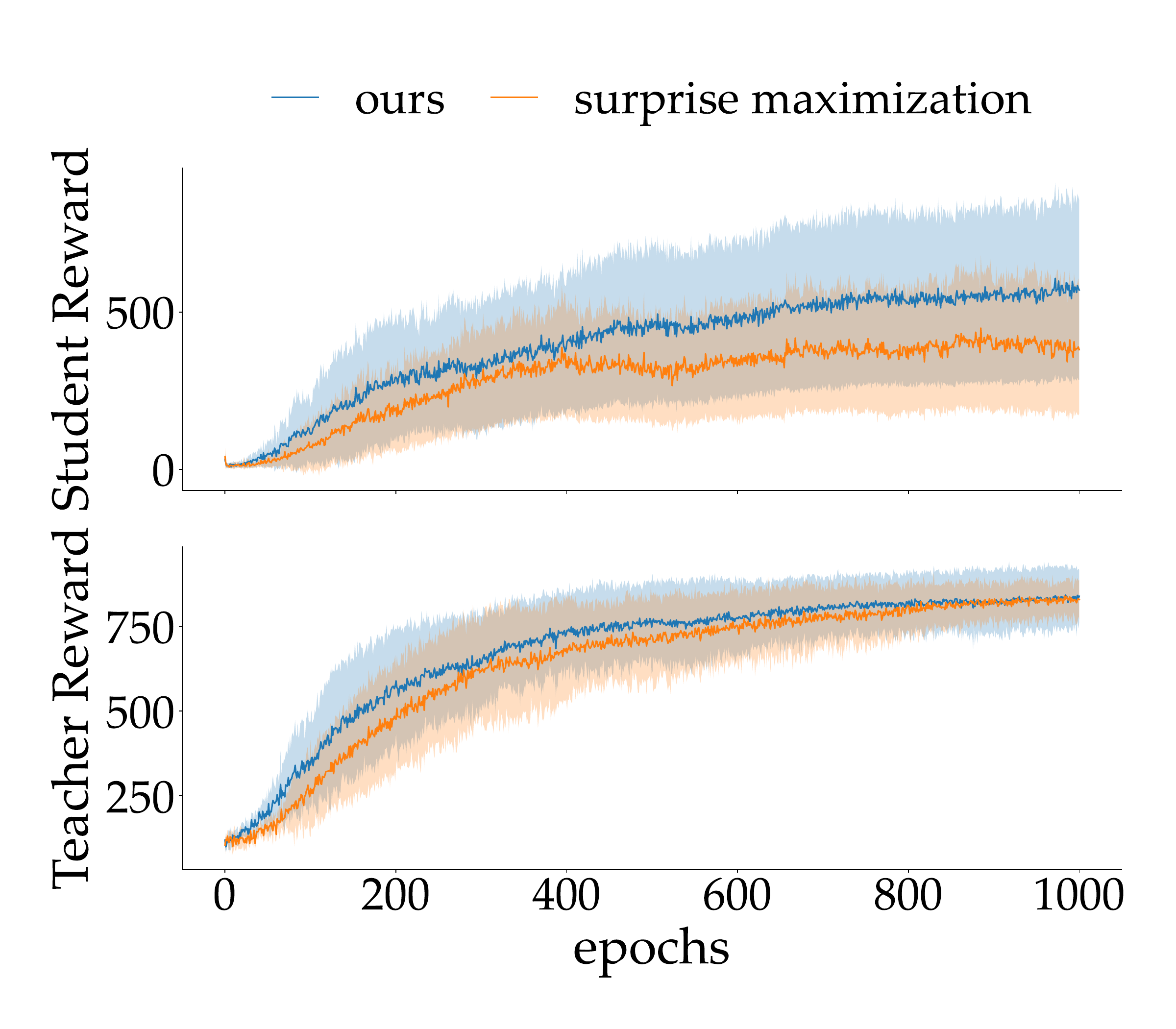}    
    \end{minipage}
    }\\
    \subfigure[CartPole SwingUp - Different $x_{pos}$ Constraint]{
    \begin{minipage}{0.48\textwidth}
        \centering
        \includegraphics[width=0.75\linewidth]{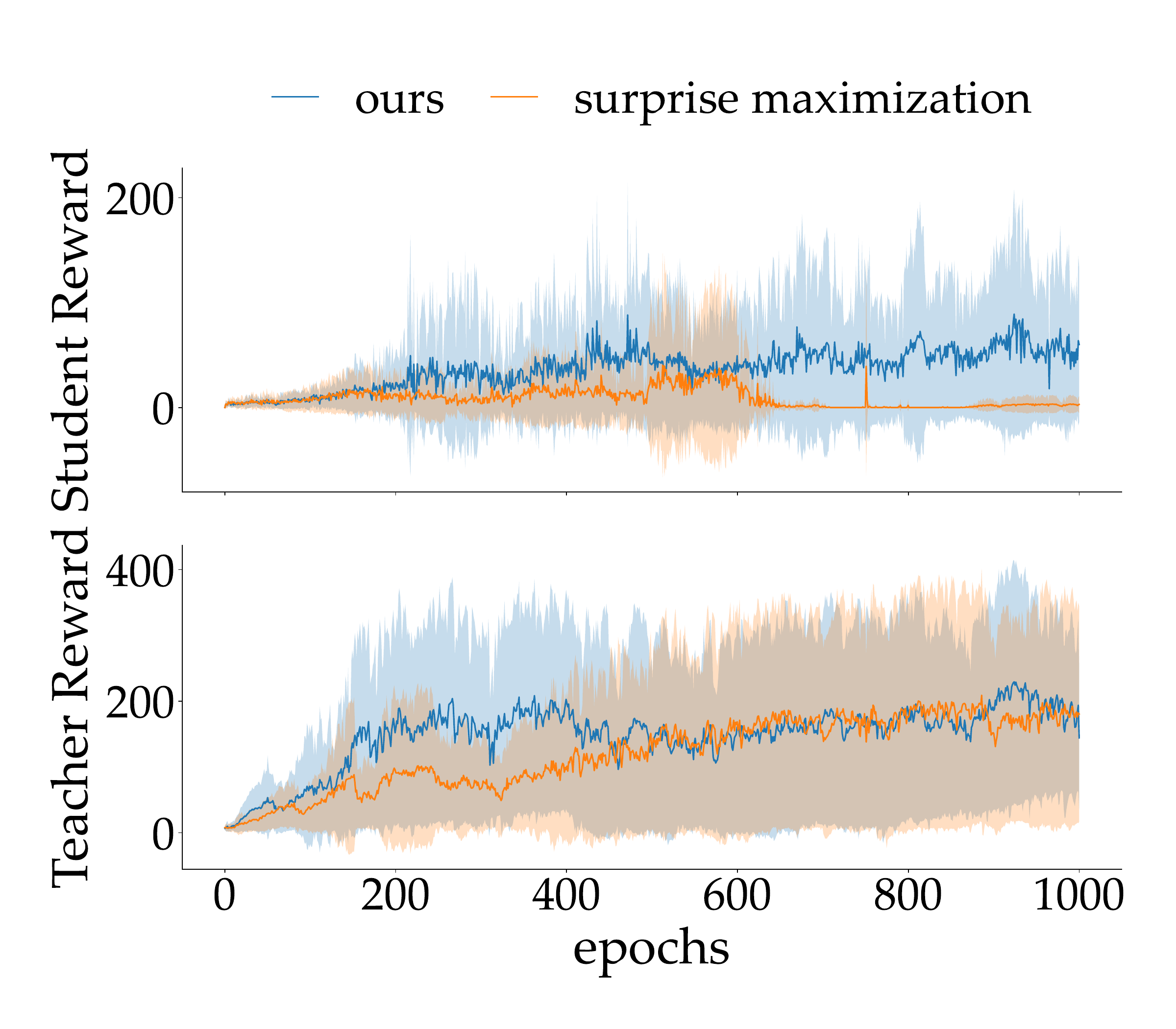}
    \end{minipage}
    }
    %\hspace{1cm}
    \subfigure[CartPole SwingUp - Different Pole Mass]{
    \begin{minipage}{0.48\textwidth}
        \centering
        \includegraphics[width=0.75\linewidth]{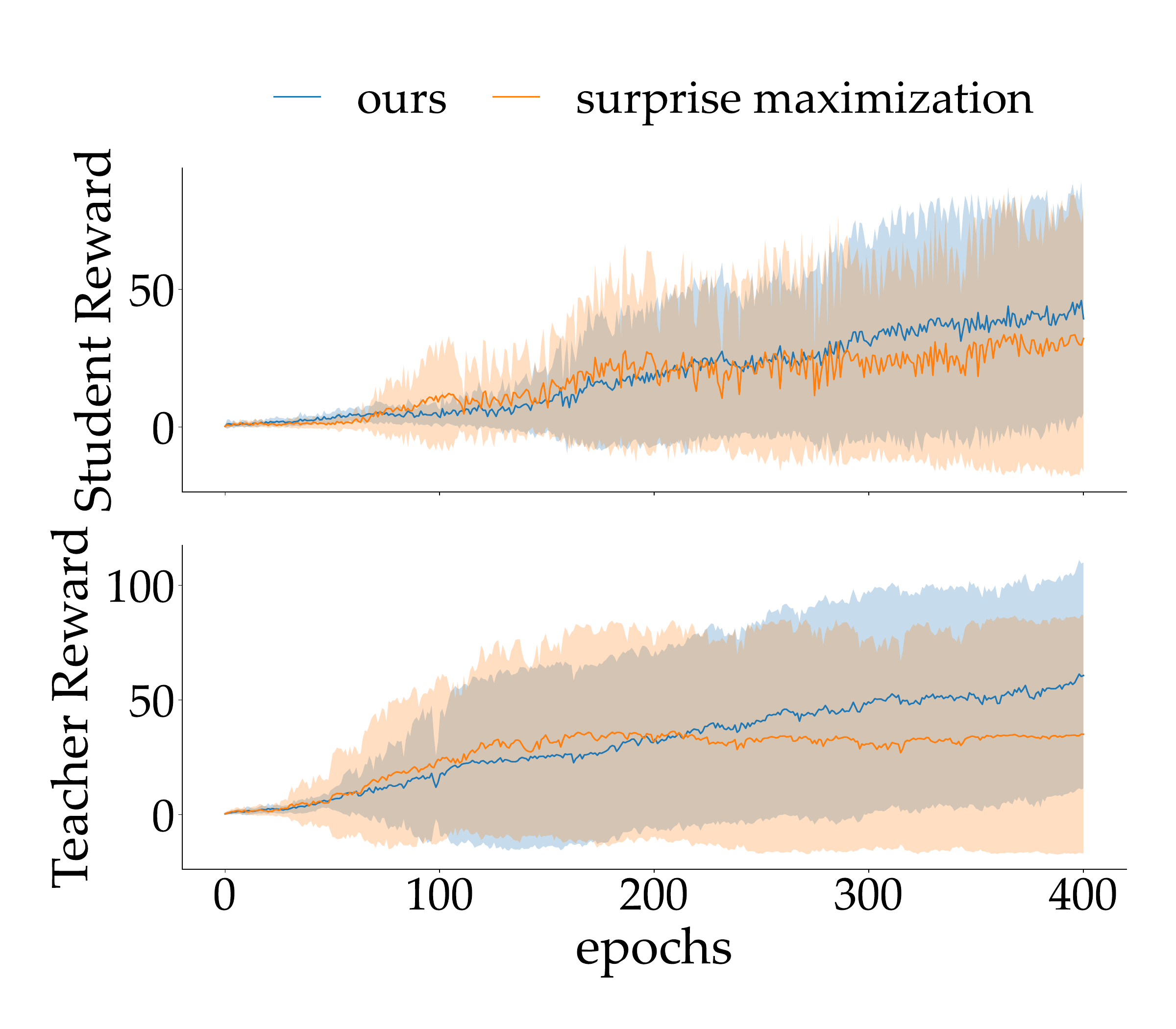}
    \end{minipage}
    }
    \caption{\small
    Teacher and Student training results where each agent has different constraints or dynamics. While the average reward of the Teacher is similar for both methods, the Student learning from our Teacher achieves higher average rewards. These show that our method can provide better demonstrations for the Student with different constraints/dynamics.
    }
    \label{fig:diff_env_lc}
    \vspace{-3mm}
\end{figure}

\subsubsection{Results}
Figure~\ref{fig:diff_env_lc} presents the training results of each experiment. In contrast to Figure~\ref{fig:std_env_lc}, where the Student's performance closely mirrors that of the Teacher, Figure~\ref{fig:diff_env_lc} exhibits a notable gap between the returns of the Teacher and the Student due to differences in their environments. Furthermore, the Student's average return is higher in our method compared to the baseline, with the exception of the Mountain Car environment. Despite minimal differences in the Teacher's performance across the two methods, there is a noticeable disparity in the Student's rewards. This observation challenges the expectation that the Student’s behavior cloning performance should closely mirror that of the Teacher, indicating that our method produces trajectories more conducive to the Student's learning while maintaining effective learning of the Teacher. Therefore, we can conclude that the Teacher's objective of minimizing the Student's surprise can improve the Student's performance, even in the absence of explicit knowledge about these discrepancies.

\begin{figure}[h]
    \centering
    %\subfigure[Epoch 0]{\includegraphics[width  = 0.29\textwidth]{figures/diff_env/mc/mc_ep0_expl.pdf}}{}
    %\subfigure[Epoch 100]{\includegraphics[width  = 0.29\textwidth]{figures/diff_env/mc/mc_ep100_expl.pdf}}{}
    %\subfigure[Epoch 200]{\includegraphics[width  = 0.29\textwidth]{figures/diff_env/mc/mc_ep200_expl.pdf}}{}
    \includegraphics[width = .7\textwidth]{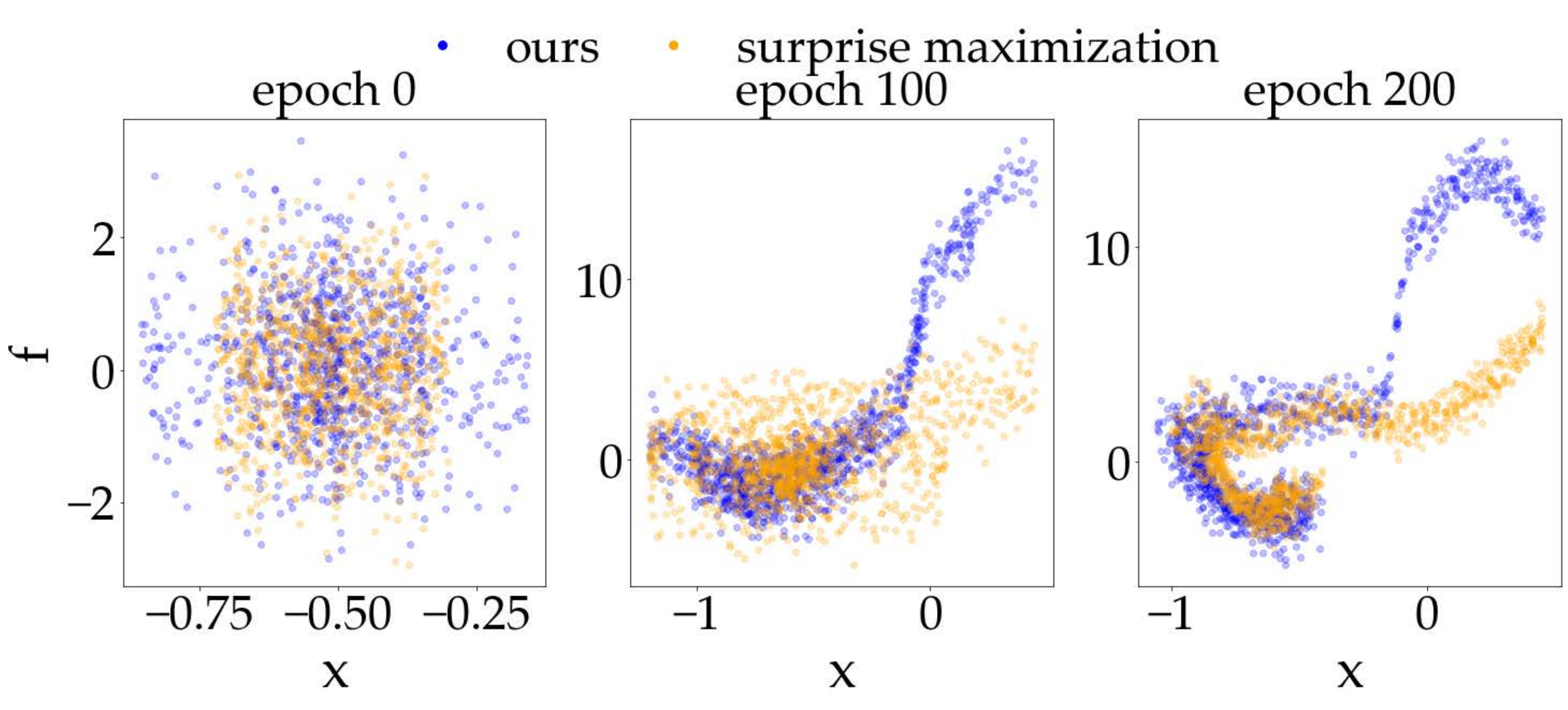}
    \caption{\small
    Teacher demonstration for Mountain Car environment where the Student has less power available than the Teacher. In training epoch 0, both methods appear to be similarly random. In the 100th epoch, our method begins to exhibit larger forces corresponding to the f-axis on the figures. At the end of training, we see there is a clear distinction between the forces exhibited by the two methods. Our method adapts to the low-power dynamics of the Student environment by demonstrating much larger forces compared to the surprise maximization algorithm.
    }
    \label{fig:mc_expl}
    %\vspace{-3mm}
\end{figure}

% In the Mountain Car environment \textcolor{red}{we see in Figure \ref{fig:diff_env_lc}, either method is able to learn an optimal policy in a simple environment such as this, even with differing strategies. The Teacher in our method has somewhat slower learning than the other method, most likely because it is simultaneously learning to solve the environment and adapt to the Student's needs.} 
In Figure~\ref{fig:mc_expl}, we further analyze the Teacher's behavior by plotting its demonstration trajectories throughout training in the Mountain Car environment. Initially, the trajectories of each algorithm appear similar. However, as training progresses, significant differences emerge in the actions taken at specific points in the state space. By the end of training, it is evident that the Teacher using our method has adapted to demonstrate larger forces. This adaptation is particularly beneficial for the Student agent's learning process, as it aligns better with the Student's need to apply greater force due to its lower power configuration.

\begin{figure}[h]
    \centering
    \subfigure[$\eta_{0_T}=0.001, \eta_{0_S}=0.001$]{
    \begin{minipage}{0.48\textwidth}
        \centering
        \includegraphics[width=0.75\linewidth]{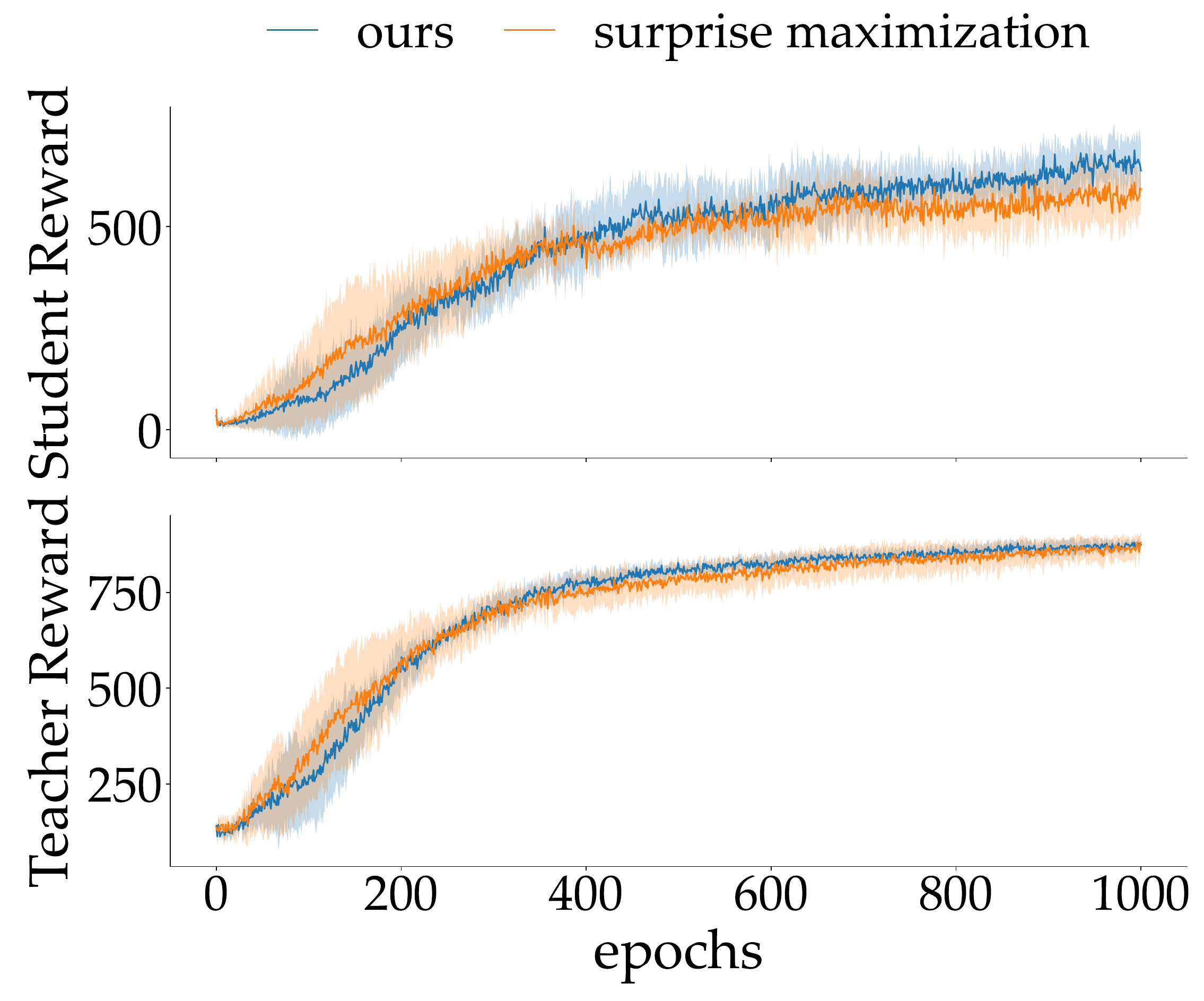}
    \end{minipage}
    }
    \subfigure[$\eta_{0_T}=0.001, \eta_{0_S}=0.005$]{
    \begin{minipage}{0.48\textwidth}
        \centering
        \includegraphics[width=0.75\linewidth]{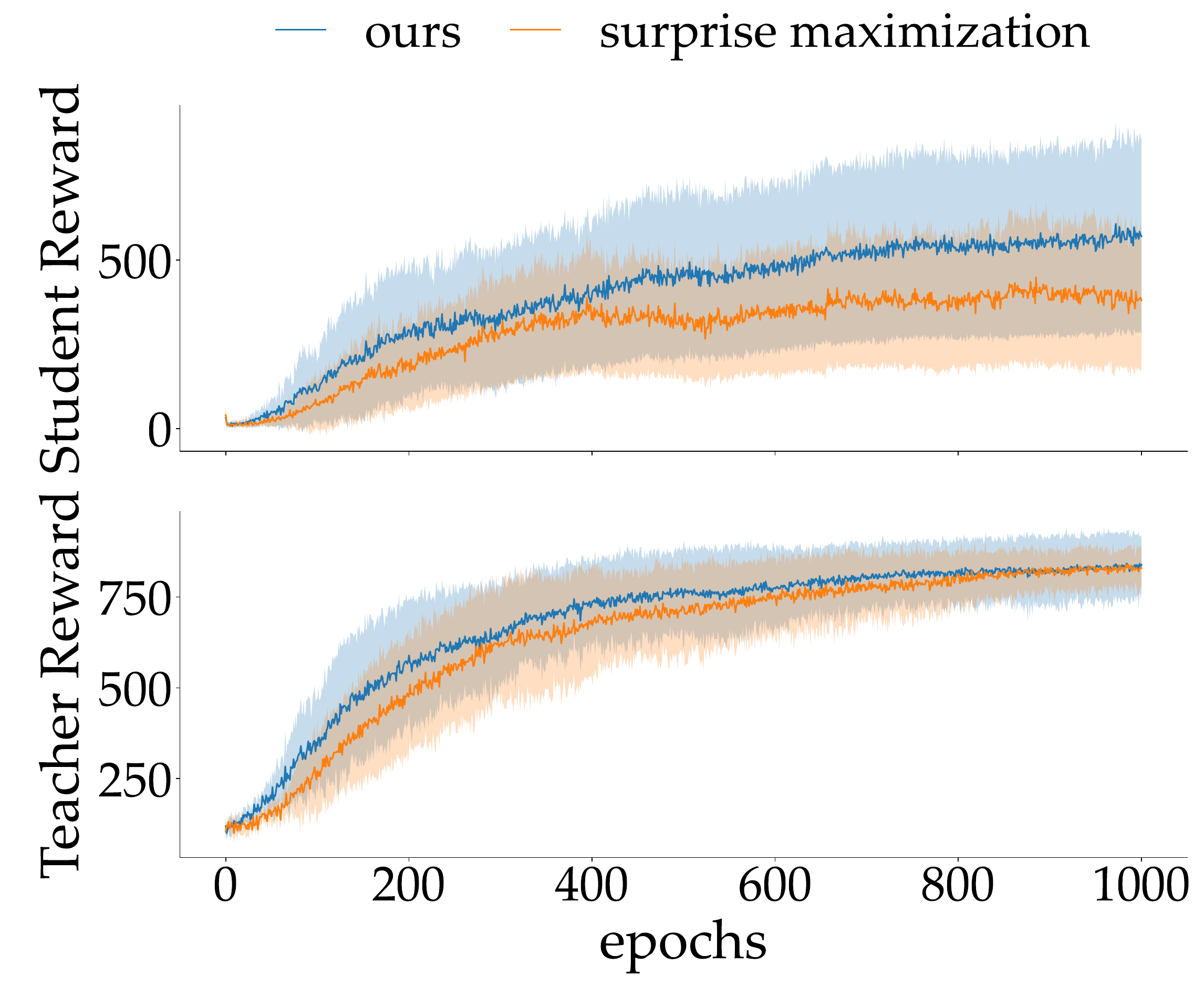}
    \end{minipage}
    }
    \caption{\small
    Training results in a sparse Half Cheetah environment with varying weights on Student surprise. The performance gap of the Student widens with an increased weight on Student surprise. This suggests that placing greater emphasis on Student surprise leads the Teacher to provide demonstrations that are more easily followed by the Student.
    }
    \label{fig:hc_diff_weight}
    \vspace{-7mm}
\end{figure}

We further investigate the impact of the Student surprise term by varying its weight in the sparse Half Cheetah environment. Figure~\ref{fig:hc_diff_weight} reveals that a higher emphasis on the Student surprise results in an increased performance gap between Student agents learning from the two different Teachers. This suggests that prioritizing the Student's surprise in the Teacher's objective encourages the Teacher to develop a policy that is more closely aligned with the Student's capabilities, which may differ from those of the Teacher.

\section{Conclusions and Future works}

We proposed a Teacher-Student learning framework, where the Teacher adapts its policy to demonstrate pedagogically effective trajectories to a Student agent acting under different constraints or dynamics parameters. This is achieved by minimizing Student surprise with respect to the Teacher's demonstration, while simultaneously maximizing the Teacher's surprise to encourage exploration. We implemented this algorithm in sparse reward environments and demonstrated that a behavior cloning agent learns more effectively from a Teacher trained with our method. We showed that our method can adapt the teaching trajectories such that the Student learns more efficiently by examining the demonstrations of the Teacher agent in the Mountain Car environment. Moreover, our method achieved comparable performance to baseline algorithms when both the Teacher and the Student were learning within the same environment settings.

As future works, we plan to extend our algorithm to the offline-online settings where transition dynamics of agents or Teacher policy are pre-trained offline. Another interesting future work could allow the Teacher to learn latent strategies to predict the transition dynamics of the Student rather than having full access to the Student's learned transition probability functions. 
% Additionally, when the Teacher and Student have \textcolor{red}{different state space and action space}, we can use shared latent spaces between the Teacher and Student to project state-action pairs~\citep{7966379}. 

\acks{This work is supported by the National Science Foundation, under grants CNS-2423130 and CCF-2423131.
The authors would like to thank Dr. Jean-Baptiste Bouvier for his valuable feedback.}

\bibliography{references}

\end{document}